\documentclass[letterpaper]{article} % DO NOT CHANGE THIS
\usepackage[submission]{aaai24}  % DO NOT CHANGE THIS
\usepackage{times}  % DO NOT CHANGE THIS
\usepackage{helvet}  % DO NOT CHANGE THIS
\usepackage{courier}  % DO NOT CHANGE THIS
\usepackage[hyphens]{url}  % DO NOT CHANGE THIS
\usepackage{graphicx} % DO NOT CHANGE THIS
\usepackage{natbib}  % DO NOT CHANGE THIS AND DO NOT ADD ANY OPTIONS TO IT
\usepackage{caption} % DO NOT CHANGE THIS AND DO NOT ADD ANY OPTIONS TO IT
\usepackage{algorithm}
\usepackage{algorithmic}
\usepackage{multirow}
\usepackage{amsmath,amssymb}
\usepackage{xcolor}
\usepackage{enumitem}
\usepackage{booktabs}

\usepackage{lipsum}

\usepackage[switch]{lineno}

\usepackage{newfloat}
\usepackage{listings}
\DeclareCaptionStyle{ruled}{labelfont=normalfont,labelsep=colon,strut=off} % DO NOT CHANGE THIS
\floatstyle{ruled}
\newfloat{listing}{tb}{lst}{}
\floatname{listing}{Listing}
\def\Vec#1{{\boldsymbol{#1}}}
\def\Mat#1{{\boldsymbol{#1}}}
\def\etal{\textit{et al.}~}
\def\ie{\textit{i.e.,}~}
\def\eg{\textit{e.g.,}~}
\usepackage{xspace}
\makeatletter
\DeclareRobustCommand\onedot{\futurelet\@let@token\@onedot}
\def\@onedot{\ifx\@let@token.\else.\null\fi\xspace}

\usepackage{array}
\newcolumntype{C}[1]{>{\centering}m{#1}}
\newcolumntype{M}[1]{>{\centering\let\newline\\\arraybackslash\hspace{-2pt}}m{#1}}

\title{LongDanceDiff: Long-term Dance Generation with Conditional Diffusion Model}
\author{
    Siqi Yang,
    Zejun Yang,
    Zhisheng Wang
}
\affiliations{
    Tencent\\
}

\usepackage{bibentry}
\begin{document}
% \linenumbers
\pagenumbering{arabic}
% \maketitle

\twocolumn[{
\renewcommand\twocolumn[1][]{#1}
\maketitle
\vspace{-50pt}
\begin{center}
    \captionsetup{type=figure}
    \includegraphics[width=1\textwidth]{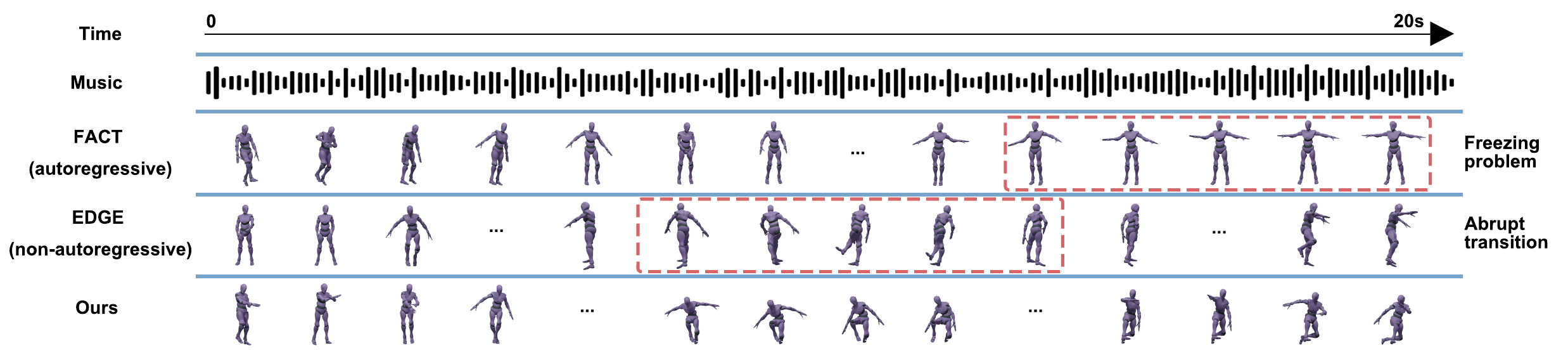}
    \captionof{figure}{Our proposed, LongDanceDiff, generates natural dance sequences with temporal coherency and spatial diversity conditioned on music. In comparison to the autoregressive method FACT~\cite{FACT}, which encounters a freezing problem, and the non-autoregressive method EDGE~\cite{tseng2022edge}, which experiences an abrupt transition issue, LongDanceDiff offers a superior solution.}
    \label{fig:frontpage}
\end{center}
}]

% \begin{figure*}[t]
% % \vspace{-5pt}
% % % \setlength{\belowdisplayskip}{0pt}
% % % \setlength{\abovedisplayskip}{0pt}
% \begin{center}
%     \includegraphics[width=1\textwidth]{figures/cover_page.png}
%     \caption{LongDanceDiff generates natural dance sequences with spatial coherence conditioned on music.}
% % % \vspace{-10pt}
% \end{center}
% % \label{fig:training_pipe_line}
% \end{figure*}

\begin{abstract}
Dancing with music is always an essential human art form to express emotion. 
Due to the high temporal-spacial complexity, long-term 3D realist dance generation synchronized with music is challenging.
Existing methods suffer from the freezing problem when generating long-term dances due to error accumulation and training-inference discrepancy.
To address this, we design a conditional diffusion model, \textbf{LongDanceDiff}, for this sequence-to-sequence long-term dance generation, addressing the challenges of temporal coherency and spatial constraint. 
LongDanceDiff contains a transformer-based diffusion model, where the input is a concatenation of music, past motions, and noised future motions. This partial noising strategy leverages the full-attention mechanism and learns the dependencies among music and past motions.
To enhance the diversity of generated dance motions and mitigate the freezing problem, we introduce a mutual information minimization objective that regularizes the dependency between past and future motions. 
We also address common visual quality issues in dance generation, such as foot sliding and unsmooth motion, by incorporating spatial constraints through a Global-Trajectory Modulation (GTM) layer and motion perceptual losses, thereby improving the smoothness and naturalness of motion generation. 
Extensive experiments demonstrate a significant improvement in our approach over the existing state-of-the-art methods.
We plan to release our codes and models soon.
\end{abstract}

\section{Introduction}

% BRIEF INTRO OF DANCE GENERATION
%
Throughout human history, music and dance have been closely intertwined.
In recent years, there has been growing interest in the development of computational methods for generating realistic 3D dance sequences that are synchronized with music.
Such methods can facilitate the creation of new dance routines or produce virtual 3D dance performances for various media, \eg video games, music videos, or virtual reality. This will result in reducing the costs associated with manual animation or optical motion capture solutions.
%
% CHALLENGES OF DANCE GENERATION
%spatial-temporal complexity
% 1. temporal
% 2. spatial
However, the generation of music-conditioned dance sequences is a challenging task that necessitates the addressing of two major aspects:
1) \textbf{temporal coherency}, which demands the synchronization of generated dance movements with the music's rhythm and beats while maintaining fluent movements and
2) \textbf{spatial constraint}, which involves generating physically feasible dance movements that adhere to the stylistic norms of dance.
Generating long-term dance (\ie 2-minute) is more challenging.

%%%%%%%%%%%%%%%%%%%%%%%%%%%%%%%
%% existing works problem
% LONG-TERM DANCE PROBLEM
% 1. freezing problem, auto regressive based method like FACT, bailando
% 2. Coherency problem, like EDGE

Existing dance generation methods can be broadly classified into two categories: auto-regressive and non-auto-regressive.
Auto-regressive methods~\cite{FACT,bailando,dance_revolution,you_never_stop}, generate future motions based on past motions. However, these methods often suffer from freezing or small-magnitude motions after several seconds, particularly when generating long-term dance sequences, as shown in Figure~\ref{fig:frontpage}.
% These issues primarily stem from the error accumulation and training-inference discrepancy of past motions.
These issues arise from error accumulation and training-inference discrepancy. 
To mitigate the training-inference discrepancy, Huang~\etal\cite{dance_revolution} proposed a curriculum strategy that alternatively feeds the predicted motions during training, which is time-consuming and does not fully resolve the problem.
Other works~\cite{bailando,you_never_stop} propose to project the dance motions to a low-dimensional manifold to constrain the distribution of generated motions.
Balilando~\cite{bailando} employed an auto-encoder with a finite dictionary of quantized dancing units by VQ-VAE\cite{vqvae}, while Sun~\etal\cite{you_never_stop} utilized the manifold bank as a refinement for noisy predictions.
However, the small number of items in the manifold bank and lack of high-fidelity training data can lead to overfitting of the motion decoder. 
This overfitting can result in low-performance motions, \eg foot-sliding, unrealistic motions.
Furthermore, the limited number of items in the manifold bank may not adequately capture the full complexity of the data, leading to a lack of diversity in the generated dance sequences.
%%%%%%%
On the other hand, non-autoregressive, such as those presented in~\cite{chen2021choreomaster,tseng2022edge}, treat each motion segment independently and use simple temporal smoothing techniques to combine the segments. But these methods may produce abrupt and unnatural transitions between poses (see Figure~\ref{fig:frontpage}).

In this paper, we focus on the long-term realistic 3D dance generation conditioned on music.
We propose a novel approach that leverages a transformer-based diffusion model conditioned on both music and past motions. 
The use of a conditional diffusion model~\cite{ho2020_DDPM,sohl2015diffusion} is motivated by its ability to generate high-quality samples with a broad range of variation by modeling the probability distribution of sequential data, making it ideal for dance generation. 
By conditioning on music and past motions, we can generate dance sequences that are not only in sync with the music but also maintain continuity with the previous movements.
Unlike text-to-motion methods~\cite{tevet2022human,zhang2022motiondiffuse} that condition at the sentence level, our approach is designed for a sequence-to-sequence setting, conditioning on frame-level music and past motions. 
Inspired by text generation~\cite{gong2022diffuseq}, we condition the diffusion process on music and past motions by concatenating them with the noised future motions as an input of a Transformer~\cite{vaswani2017attention}.
This approach allows the Transformer to model long-term dependencies among the music, past motions, and noised future motions with full attention during the reverse diffusion process, enabling the generation of diverse dance sequences closely synchronized with the music and past motions.
To further improve the model's performance, we propose a mutual information minimization objective that regularizes the dependency between past and future motions. It is based on the intuition that while past motions provide crucial context for generating future movements, an over-reliance on this information can lead to overfitting and limit the diversity of the generated sequences. By minimizing the mutual information,  the objective is to diminish this reliance and promote the diversity of the generated sequences, thereby alleviating the issue of freezing.
Our proposed method, therefore, not only takes into account the temporal coherency of dance movements but also ensures a balance between leveraging past motion information and maintaining the diversity and fluidity of the generated sequences.
Another significant challenge in dance generation is the visual quality of the generated motion. 
Common issues such as foot sliding, unsmooth motion, and unreasonable pose often compromise the realism of the generated dance sequences.
The existing transformer-based models, which neither directly acquire nor utilize spatial information, frequently overlook these problems
To address these challenges, we propose an enhanced approach that incorporates spatial constraints into the dance generation process.
Foot sliding or drifting is a common yet overlooked problem in motion and dance generation, where the feet of a virtual character appear to slide or move unnaturally across the ground, even when the rest of the body appears to be moving correctly.
It often arises from a misalignment between the global trajectory of the root joint and the local rotations of other body joints.
To mitigate this, we introduce a novel Global-Trajectory Modulation (GTM) layer to learn the interdependence between the global trajectory and the local rotations of other body joints. 
Furthermore, to enhance the smoothness and naturalness of motion generation, we employ motion perceptual losses as a form of regularization in joint position, motion velocity, and foot contact.
Together, these approaches provide a comprehensive framework for producing high-quality dance sequences with improved temporal coherence and natural-looking motion.

% % spatial attention

% The spatial attention block draws information from
% the joint features at the current time step whereas the temporal
% block focuses on distilling information from the previous
% time steps of individual joints.

% summary
 % Our results demonstrate the potential of our approach for generating compelling and synchronized dance sequences that are closely tied to musical input.

% contributions
In summary, our contributions are as follows:
% \begin{itemize}
1) We propose a conditional diffusion model for the sequence-to-sequence long-term dance generation, where the partial noising strategy learns the dependencies among music and past motions. Our method effectively can generate diverse and high-fidelity long-term dance sequences.
2) We propose a novel mutual information minimization regularizer to reduce the over-dependency on past motions, which enhances the diversity of motion and effectively mitigates the freezing problem.
3) We propose to incorporate spatial constraints to the diffusion model through a global-trajectory modulation layer to mitigate the foot sliding problem and motion perpetual losses to improve the quality of dance motions.
4) We evaluate our method on a dataset of dance sequences synchronized with music, showing that it outperforms several baseline methods in terms of both visual quality and musical synchronization.
% \end{itemize}

% \begin{figure}[t]
% % \vspace{-5pt}
% % \setlength{\belowdisplayskip}{0pt}
% % \setlength{\abovedisplayskip}{0pt}
% \begin{center}
%    \includegraphics[width=1\linewidth]{figures/training_pipe_line.pdf}
% \end{center}
% % \vspace{-10pt}
%    \caption{}
% \label{fig:training_pipe_line}
% \end{figure}

%------------------------------------------------------------------------
\section{Related Works}

\begin{figure*}[t]
% \vspace{-5pt}
% \setlength{\belowdisplayskip}{0pt}
% \setlength{\abovedisplayskip}{0pt}
\begin{center}
   \includegraphics[width=1\linewidth]{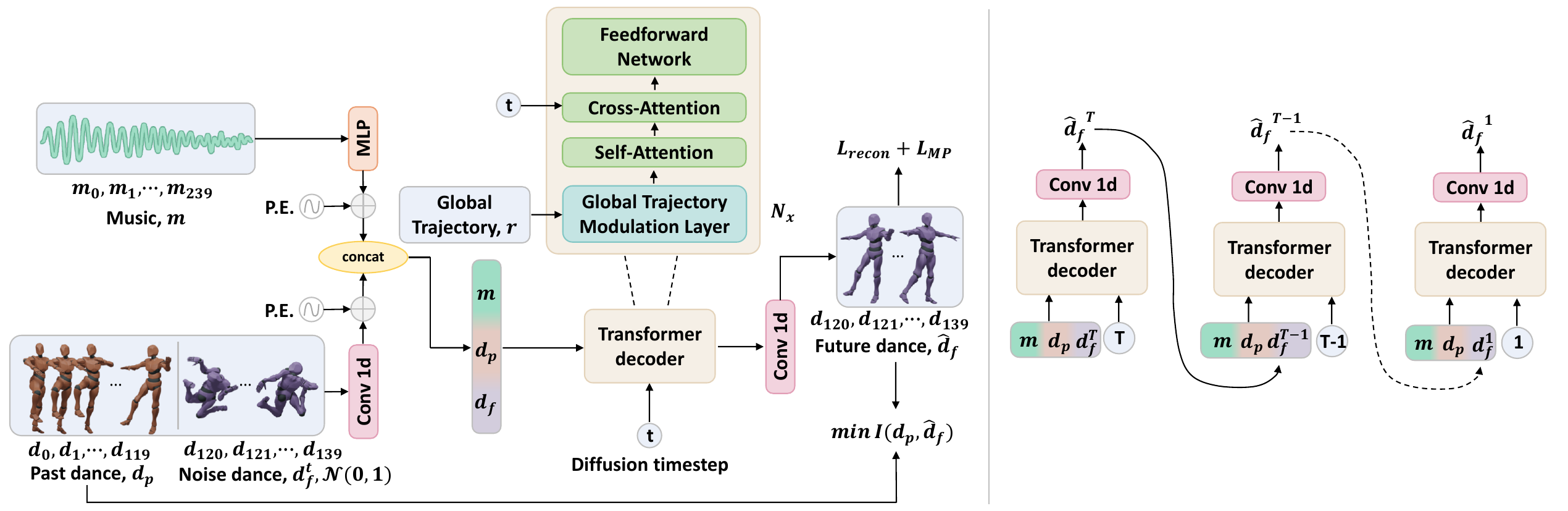}
\end{center}
% \vspace{-10pt}
   \caption{\textbf{Overview of LongDanceDiff.} LoneDanceDiff is a music and past motion-conditioned diffusion model based on Transformers. The forward passing of diffusion is with partial nosing, where the input of Transformer is a concatenation of music $\Vec{m}$ (240 frames), past motions ${\Vec{d}_p}$ (120 frames), and noised future motions ${\Vec{d}_f}^t$ (20 frames). The real past motions and noisy motions share the same embedding function for feature alignment. To alleviate the foot sliding issue, a Global-Trajectory Modulation layer is introduced, while a motion perceptual loss $\mathcal{L}_\textrm{MP}$ is proposed to learn spatial constraints. To regulate the dependency on past motions and promote motion diversity, a mutual information minimization objective is jointly optimized.}
\label{fig:framework}
\end{figure*}

%-------------------------------------------------------------------------
% refers to never stop
\subsection{Dance Generation}
% The field of music-conditioned dance generation has seen significant growth in recent years. 
% retrieval based
% Several earliest works in the music-conditioned dance generation~\cite{ofli2008audio,shiratori2006dancing,fan2011example,ofli2011learn2dance} have employed retrieval-based methods to generate complete dance sequences.
Early studies in the music-conditioned dance generation~\cite{ofli2008audio,shiratori2006dancing,fan2011example,ofli2011learn2dance} have employed retrieval-based methods to generate complete dance sequences.
These approaches involve selecting pre-defined motion segments from a pre-existing database based on the music, and arranging them in a sequence with appropriate transitions. 
Chen~\etal\cite{chen2021choreomaster} traverses the node transition routines on a motion graph and introduces choreography-oriented constraints to compose the final dance motion sequences.
However, this approach requires significant human effort to determine motion segments, and is limited to generating motions within the dataset. Additionally, the transitions between dance segments are often unrealistic.
% prediction based
More recent works treat dance generation as a motion prediction problem and synthesize dance from scratch. A number of network architectures have been proposed, including CNNs~\cite{kritsis2022danceconv, holden2016deep}, RNNs~\cite{huang2020dance, tang2018dance, alemi2017groovenet, yalta2019weakly}, GCNs~\cite{ren2020self, ferreira2021learning, yan2019convolutional}, GANs~\cite{sun2020deepdance, lee2019dancing} and Transformers~\cite{li2022danceformer, li2021ai, huang2022genre}.
% Bailando
Li~\etal\cite{bailando} proposed the Bailando, which contains a choreographic memory to learn meaningful dancing units and an actor-critic generative pre-trained transformer to compose dancing units to fluent dance coherent to the music. 
% Consequently, Bailando can create long rhythmic dance sequences in a uniform style.
% Never stop dance
Since the high-dimensional motion data are vulnerable to noise, Sun~\etal\cite{you_never_stop} considered that learning a low-dimensional manifold representation can reduce the influence of noise. 
% They provided two flexible modules, which can be easily plugged into other existing methods, to achieve non-freezing large-magnitude motion generation.
% Dance revolution
Huang~\etal\cite{dance_revolution} proposed a novel curriculum learning strategy is utilized to alleviate error accumulation when generating long motion sequences.
However, curriculum learning needs a large amount of computations and their performance is still restricted.
Recently, Tseng~\etal\cite{tseng2022edge} have combined a strong music feature extractor, Jukebox, and a transformer-based diffusion model to generate editable dance from music. However, this method does not explicitly consider past motions and is not ideal for generating choreographies with very long-term dependencies.

%-------------------------------------------------------------------------
% refer to EDGE
\subsection{Generative Diffusion Methods}
%general diffusion
% Diffusion models~\cite{ho2020_DDPM,sohl2015diffusion} belong to the class of deep generative models that rely on the stochastic diffusion process.
Diffusion models~\cite{ho2020_DDPM,sohl2015diffusion} are deep generative models that use stochastic diffusion processes.
% Specifically, a sample from the data distribution is gradually noised by the diffusion process.
% Conversely, in the reverse process, a model gradually removes noise from a sample that has been generated by a simple distribution (such as Gaussian noise) and learns to map the data distribution.
% In the past few years, diffusion-based methods have been applied to a wide range of tasks and shown superior performance of generative diversity and facticity, including text-to-image generation\cite{rombach2021highresolution, ruiz2022dreambooth, gu2022vector}, music generation\cite{mittal2021symbolic, hawthorne2022multi, schneider2023mo}, seq2seq generation\cite{gong2022diffuseq, yuan2022seqdiffuseq} and text-to-motion generation\cite{ren2022diffusion, zhang2022motiondiffuse, tevet2022human, zhao2023modiff}...
Diffusion-based methods have been applied to various tasks and shown superior performance of generative diversity and facticity, including text-to-image generation\cite{rombach2021highresolution, ruiz2022dreambooth, gu2022vector}, music generation\cite{mittal2021symbolic, hawthorne2022multi, schneider2023mo}, seq2seq generation\cite{gong2022diffuseq, yuan2022seqdiffuseq} and text-to-motion generation\cite{ren2022diffusion, zhang2022motiondiffuse, tevet2022human, zhao2023modiff}...
% conditioned generation, classifier-guided, classifier-free, seq2seq
As for conditional generation, Dhariwal and Nichol~\cite{diffusion_beat_gan} introduced classifier-guided diffusion. 
The Classifier-Free Guidance approach proposed by Ho and Salimans~\cite{ho2022classifier_free_guid} enables conditioning while trading-off fidelity and diversity. 
% Furthermore, Stable Diffusion\cite{rombach2021highresolution} has shown its powerful capability of high-quality image synthesis from text. 
Stable Diffusion\cite{rombach2021highresolution} has also shown its ability to synthesize high-quality images from text.
% Stable Diffusion only relies on several prompts to create diverse images relative to prompts, significantly improving human productivity.
% 
% text to motion
Recently, diffusion-based methods have been studied in text-to-motion tasks~\cite{tevet2022human,zhang2022motiondiffuse}.
% Despite the fact that text-to-motion and music-conditioned dance generation tasks exhibit comparable high-level features, the latter is more demanding due to its reliance on sequential data conditioning. 
The text conditions are usually attributes that are easy to model, whereas
music and past motions as conditions in the sequence-to-sequence problem of dance generation are much more complex.

% % dance
% More recently, concurrent with this work, Tseng~\etal\cite{tseng2022edge} have suggested
% diffusion models for dance generation. 
% However, their proposed methods only focus on generating a 5-second dance motion clip without considering motion context.
% And therefore, it is not ideal for long-term dance generation.
% Our proposed method differs from theirs in incorporating previous motion as one of the conditions and making an effective design for conditional dance generation. 
% % seq2seq diffusion

% Compared with our work, we focus on the SEQ2SEQ diffusion
% models for text generation in the continuous space and our work is the first to explore this
% setting to the best of our knowledge.

%------------------------------------------------------------------------
\section{Methodology}

% \begin{figure*}[t]
% % \vspace{-5pt}
% % \setlength{\belowdisplayskip}{0pt}
% % \setlength{\abovedisplayskip}{0pt}
% \begin{center}
%    \includegraphics[width=1\linewidth]{figures/framework.pdf}
% \end{center}
% % \vspace{-10pt}
%    \caption{}
% \label{fig:framework}
% \end{figure*}

% Our goal is to design a new image synthesis paradigm
% utilizing parallel decoding and bi-directional generation.
% We follow the two-stage recipe discussed in 2.1, as illustrated
% in Figure 3. Since our goal is to improve the second
% stage, we employ the same setup for the first stage as in the
% VQGAN model [15], and leave potential improvements to
% the tokenization step to future work.
% For the second stage, we propose to learn a bidirectional
% transformer by Masked Visual Token Modeling (MVTM).
% We introduce MVTM training in 3.1 and the sampling procedure
% in 3.2. We then

% Our goal is to design a model that can generate long-term realistic dance sequences with high fidelity and diversity.
We propose a conditional diffusion model that is conditioned by music and past motions, shown in Figure~\ref{fig:framework}.
% We employ a 
We introduce the details of the music-conditioned diffusion model in Sec.~\ref{sec:diffusion}, mutual information minimization in Sec.~\ref{sec: mutual info}\ and spatial constraints in Sec.~\ref{sec: spatial constraint}.

\subsection{Problem Formulation}
The goal of the music-conditioned 3D dance generation is to generate a sequence of dance motions $\mathcal{D} = \{\Vec{d}_i\}^{N}_{i=1}$ that are synchronized to a given music clip $\mathcal{M} = \{\Vec{m}_i\}^{N}_{i=1}$.
Here, $N$ is the number of frames of the music clip with a specific sampling rate. 
The dance generation process is often guided by a 2-second past motion $\mathcal{D}_p = (\Vec{d}_{1}, ..., \Vec{d}_{K})$.
Then the problem is to generate a sequence of future motion $\mathcal{D}_f = (\Vec{d}_{K+1}, ..., \Vec{d}_{N})$.
We formulate it as a conditional dance motion prediction problem.

\subsection{Music and Past Motion Conditioned Diffusion}
\label{sec:diffusion}

In this work, we propose LongDanceDiff to extend vanilla diffusion models to learn music-conditioned long-term dance generation.

% introduction to diffusion model
\medskip
\noindent
\textbf{Diffusion.}
DDPM~\cite{ho2020_DDPM} defines diffusion as a Markov noising process with T steps, $\{\Vec{x}_t \}^T_1$.
Given a data point sampled from a real-world data distribution $\Vec{x}_0$, the forward process gradually corrupts $\Vec{x}_0$ to a standard Gaussian noise.
For each forward step $t \in [1, 2, ..., T]$, the noising process is defined as:
\setlength{\belowdisplayskip}{3pt}
\setlength{\abovedisplayskip}{2pt}
\begin{equation}
    q(\Vec{x}_t|\Vec{x}_{t-1}) = \mathcal{N}(\sqrt{\alpha_t} \Vec{x}_{t-1}, (1-\alpha_t) I) \textrm{ ,}
\end{equation}
where $\alpha_t \in (0,1)$ are constant hyper-parameters that decrease monotonically.
When $\alpha_t$ approaches 0, we can approximate $\Vec{x}_T \sim \mathcal{N}(0, I)$. 
The vanilla diffusion model is unconditional and adapted to the sequence-to-sequence dance generation problem, which is conditioned by music and past motions.

% In this work, we adapt the diffusion model to the music-conditioned dance generation.

\medskip
\noindent
\textbf{Forward Passing with Partial Noising.}
% We denote ${\Vec{d}_f}^t$ as the future dance motion at noising step t.
For each forward step t, we concatenate the embeddings of music $\Vec{m}$, past motions $\Vec{d}_p$, and noised future motions ${\Vec{d}_f}^t$ to one vector.
Specifically, we gradually inject noise into the future dance $\Vec{d}_f$ and we get the input $\Vec{z}^t$:
\begin{equation}
    \Vec{z}^t = \Vec{m} \oplus \Vec{d}_p \oplus {\Vec{d}_f}^t \textrm{ .}
\end{equation}
% where the $\Vec{d}_f^t$ is the noised future motion.
Unlike conventional diffusion models that corrupt the entire input vector of the network without distinction, we only impose noising on $\Vec{d}_f$.
This partial noising allows us to compose a cross-model embedding to the Transformer.
In this way, the attention layers with full attention in the Transformer could learn the cross-model dependencies among music, past motions, and future motions.
% The process of diffusion with partial noising is shown in Figure~\ref{fig:training_pipe_line}.
% For simplicity, we refer to $\Vec{d}_f$ as $\Vec{x}$ in the following text.

\begin{figure}[t]
% \vspace{-5pt}
% \setlength{\belowdisplayskip}{0pt}
% \setlength{\abovedisplayskip}{0pt}
\begin{center}
   \includegraphics[width=0.7\linewidth]{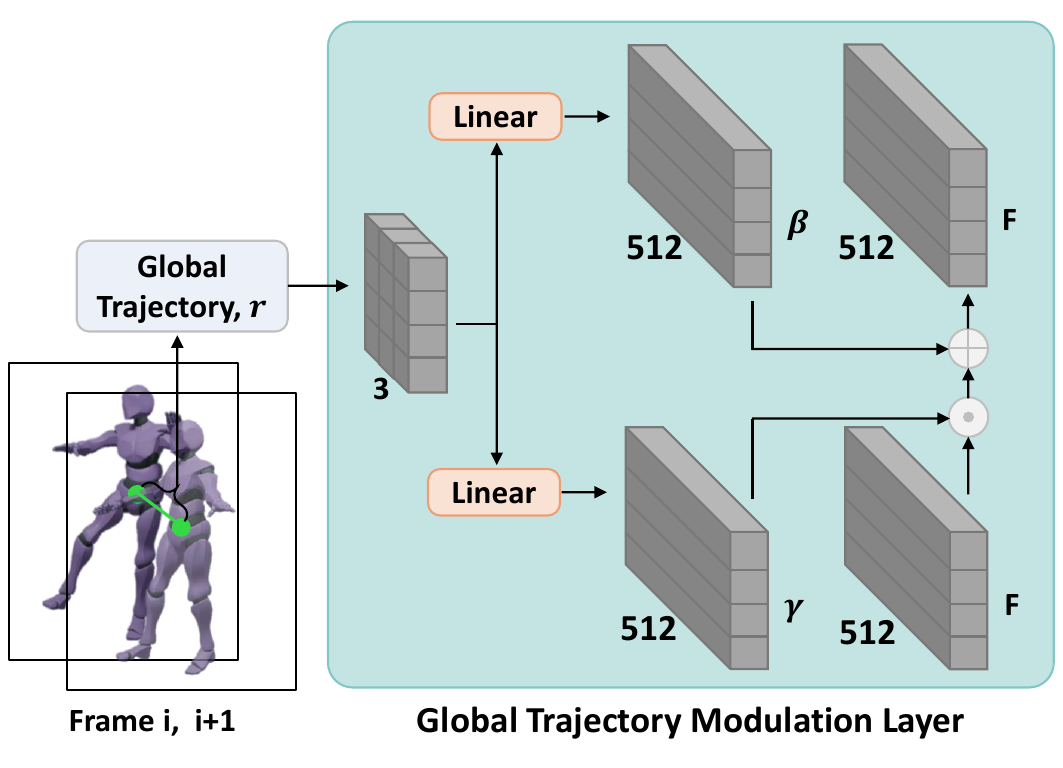}
\end{center}
\vspace{-10pt}
   \caption{The Global Trajectory Modulation (GTM) layer models the dependence between the global trajectory and the rotations of other body joints in order to mitigate the issue of foot sliding.}
\vspace{-10pt}
\label{fig:GTM}
\end{figure}

\medskip
\noindent
\textbf{Reverse Process with Conditional Denoising.}
In our work, the music-conditioned motion generation models the distribution based on the music $\Vec{m}$, past motions $\Vec{d}_p$.
Instead of predicting $\epsilon_{t}$ introduced by Ho~\etal\cite{ho2020_DDPM}, we follow Ramesh~\etal\cite{ramesh2022hierarchical} and predict the signal itself, \ie future motions $\hat{\Vec{d}}_f$.
The motivation is that we could directly regularize the predicted motions with motion perceptual constraints during the reverse process.
This objective meets the 'simple' objective function formulated by Ho~\etal\cite{ho2020_DDPM}. And we formulate it as follows:
\begin{equation}
\begin{aligned}
    \mathcal{L}_\textrm{recon} &= \mathbb{E}_{d_f \sim q(d_f |c) } [\| \Vec{d}_f - \hat{\Vec{d}}_f\|^2_2] \textrm{ ,} \\
    \textrm{where} \quad \hat{\Vec{d}}_f &= f_{\theta} ({\Vec{d}_f}^t, t, \Vec{m}, \Vec{d}_p) \textrm{ ,}
\label{reconstruction loss}
\end{aligned}    
\end{equation}
and $f_{\theta}(\cdot)$ is a Transformer decoder.
Due to the attention mechanism in the Transformer, the reconstruction of $\Vec{d}_f$ also takes $\Vec{m}$ and $\Vec{d}_p$ into account and thus affects the learning of $\Vec{d}_f$.
The attention layer is the core component in Transformer that determines the dependencies among sequential elements, \ie the concatenation of $\Vec{m}$, $\Vec{d}_p$, and $\Vec{d}_f$.
% It is implemented as
% \begin{equation}
%     \textrm{Att}(\Mat{Q},\Mat{K},\Mat{V},\Mat{M}) = \textrm{softmax}(\frac{\Mat{Q}\Mat{K}^T + \Mat{M}}{\sqrt{C}}) \Mat{V}  \textrm{ ,}
% \end{equation}
% where $\Mat{Q}$,$\Mat{K}$,$\Mat{V}$ are the query, key and value from the input $\Vec{z}^t$ in self-attention, and $\Mat{M}$ is the mask. For the diffusion timestep t, we feed it to the cross-attention module for guiding the denoising process.

We also note that we propose to use the same embedding function for both the past motions and noised future motions, which is a Conv1d layer in this work.
It enables the joint training of features of real motions and noisy motions, thereby aligning their distributions.
As a result, the training-inference discrepancy is minimized, and the issue of error accumulation leading to the freezing problem is mitigated.
The network architecture is shown in Figure~\ref{fig:framework}.

\subsection{Minimizing Mutual Information between Past and Future Motion}
\label{sec: mutual info}
Dance is inherently temporal and highly dependent on both past motions and music.
However, the freezing problem in auto-regressive methods occurs when the past motions generated from the previous step are "noisy", and are not aligned with the distribution of training data.
Diverse yet natural-looking generated motions might also be "noisy" past motion due to the overfitting of training.
Thus, it is crucial to improve the generalization of the motion generator and encourage diversity.
To this end, we propose a mutual information objective that aims to regularize the dependency between past motions and current motions. 
The underlying intuition is that the future motion of a human should not be overly deterministic or predictable based on past motion.
By minimizing the mutual information, we aim to ensure that the generated motion sequences are not merely a replication of past movements, but rather, they exhibit a degree of novelty and unpredictability, akin to natural human motion. This approach also helps to avoid overfitting to the training data, thereby enhancing the generalizability of our model. 
Combing this method with the reconstruction loss in diffusion (Eq.~\ref{reconstruction loss}), it therefore provides a balance between the coherence of the generated motion and the diversity.
% Consequently, we can minimize the distribution difference between current noisy motions and past real motions to obtain realistic motions, and also model the temporal dependency to improve the coherence in motion movements and styles.
% 

Formally, mutual information (MI) is a measure of the dependency between two random variables:
\begin{equation}
    \mathcal{I}(\Vec{v}_{1} ; \Vec{v}_2) = \mathbb{E}_{p(\Vec{v}_{1}, \Vec{v}_{2})}  \biggr[ \, \log \frac{p(\Vec{v}_{1}, \Vec{v}_{2})}{p(\Vec{v}_{1})p(\Vec{v}_{2})} \biggr] \textrm{ ,}
\end{equation}
where $p(\Vec{v}_{1}, \Vec{v}_{2})$ is the joint probability distribution between $\Vec{v}_{1}$ and $\Vec{v}_{2}$, and $p(\Vec{v}_{1})$ and $p(\Vec{v}_{2})$ are the marginal distributions respectively.
Our objective is to minimize the MI between the past motions and the predicted future motions, which can be formulated as:
\begin{equation}
\min \, \mathcal{I}(\Vec{d}_{p} ; \Vec{d}_f) \textrm{ ,}
\end{equation}
where $\mathcal{I}(\Vec{d}_{p} ; \Vec{d}_f)$ indicates the dependency between the past motions $\Vec{d}_{p}$ and future motions $\Vec{d}_{f}$.
We adopt Variational Self-Distillation (VSD)~\cite{tian2021farewell} to compute the MI:
\begin{equation}
\mathcal{L}_\textrm{MI} = - D_{KL}(\Vec{d}_{p}||\Vec{d}_{f})\textrm{ .}
\end{equation}

\subsection{Learning Spatial Constraints}
\label{sec: spatial constraint}
A high-quality, generated dance motion is required to look smooth and natural. However, there are several visual challenges in the field of dance generation, for instance, foot sliding, unsmooth motion, and unreasonable pose. Most of these problems can be formulated in the spatial dimension.
Introducing some prior knowledge may constrain the dance motion to some extent. 
However, the transformer-based diffusion model neither directly acquires nor utilizes spatial information.
Consequently, it is necessary for the diffusion model to learn practical spatial constraints to improve the rationality of the generated dance motion.

\subsubsection{Global-Trajectory Modulation Layer for Mitigating Foot-Sliding Problem.}
Foot sliding or drifting is a common yet overlooked problem in motion and dance generation, where the feet of a virtual character appear to slide or move unnaturally across the ground, even when the rest of the body appears to be moving correctly.
This problem often arises from a misalignment between the global trajectory of the root joint and the local rotations of other body joints.
While current motion generation methods could learn a probabilistic distribution of the data and generate diverse motions, they do not account for the interdependence between the global root trajectory and the local rotations of other body joints.
As a result, explicit modeling of this relationship is necessary to overcome this limitation.
To address this issue, we propose a novel approach that utilizes a modulation mechanism to adjust the latent codes of motion representations based on the global root trajectory, thereby enhancing the relationship between the two variables. 
Inspired by the FiLM layer proposed by Perez\etal\cite{perez2018film}, we propose a Global-Trajectory Modulation (GTM) layer to adaptively influence the intermediate features by applying an affine transformation. 
The GPM layer learns linear functions $f_\gamma$ and $h_\beta$ with the input of global trajectory $\{\Vec{r}_j\}^{j=3}$ of future motions:
\begin{equation}
    \gamma_{j, c} = f_\gamma(\Vec{r}_j),  \beta_{j, c} = h_\beta(\Vec{r}_j) \textrm{ ,}
\end{equation}
where $c$ is the index of feature maps.
The outputs $\gamma$ and $\beta$ modulate the per-feature-map distribution of activations via a feature-wise affine transformation:
\begin{equation}
    \Mat{F}_{j, c} = \gamma \Mat{F}_{j, c} + \beta \textrm{ .}
\end{equation}
Details of the GTM layer are illustrated in Figure~\ref{fig:GTM}.

\subsubsection{Motion Perceptual Losses}
In motion generation, networks~\cite{tevet2022human} are regularized by geometric losses to learn the physically plausible and coherent motion.
We adopt the three common geometric losses introduced by Tevet~\etal\cite{tevet2022human} to regulate (1) positions, (2) velocities, and (3) foot contacts.
\begin{align}
\footnotesize
\begin{split}
    & \mathcal{L}_\textrm{pos}  = \frac{1}{N} \sum^N_{i=1} \| FK(\Vec{d}_f^i) - FK(\hat{\Vec{d}_f}^i) \|^2_2 \textrm{ ,} 
\end{split}
\\
\footnotesize
\begin{split}
    & \mathcal{L}_\textrm{vel} = \frac{1}{N-1} \sum^{N-1}_{i=1} \| (\Vec{d}_f^{i+1} - \Vec{d}_f^i) -  (\hat{\Vec{d}_f}^{i+1} - \hat{\Vec{d}_f}^i) \|^2_2 \textrm{ ,} 
\end{split}
\\
\footnotesize
\begin{split}
\footnotesize
    & \mathcal{L}_\textrm{contact} = \frac{1}{N-1} \sum^{N-1}_{i=1} \| (FK(\Vec{d}_f^{i+1}) - FK(\hat{\Vec{d}_f}^i)) \cdot f_i \|^2_2 \textrm{ ,} 
\end{split}
\end{align}
where $FK(\cdot)$ is the forward kinematic function that calculates the positions of joints given the angles of each joint, and $f_i \in \{0, 1\}$ is the binary foot contact mask for the $i$th frame.
Based on the ground truth data, the foot contact label $f_i$ is identified as 1 when the joint related to the foot is in proximity to the ground and its velocity is approximately zero, otherwise, it is 0.
The position and velocity loss encourage realistic movements that follow natural joint angles and velocities and help to prevent unrealistic movements.
The foot contact loss can eliminate the foot-sliding problem and improve the accuracy of foot-ground interactions in motion generation.
Then the motion perceptual loss is $\mathcal{L}_\textrm{MP} = \lambda_\textrm{pos} \mathcal{L}_\textrm{pos} + \lambda_\textrm{vel} \mathcal{L}_\textrm{vel} + \lambda_\textrm{contact} \mathcal{L}_\textrm{contact}$, where the $\lambda_\textrm{pos}, \lambda_\textrm{vel}, \lambda_\textrm{contact}, \lambda_\textrm{MI}$ are the hyper-parameters.

\subsection{Training Objective}
The overall training loss $\mathcal{L}$ of our proposed LongDanceDiff method is as follows:
\begin{equation}
    \mathcal{L} = \mathcal{L}_\textrm{recon} + \lambda_\textrm{MI} \mathcal{L}_\textrm{MI} + \lambda_\textrm{MP} \mathcal{L}_\textrm{MP} \textrm{ ,}
\end{equation}
where the $\lambda_\textrm{MP}, \lambda_\textrm{MI}$ are the hyper-parameters to balance the trade-off among different losses.

% \subsection{Prior}

% \subsubsection{Seq2Seq Diffuse}

% \subsubsection{Prior Sampling}

%------------------------------------------------------------------------
\section{Experiments}
This section presents an evaluation of the proposed approach for generating dance generation. 
% We first provide the implementation details of our experiments and then demonstrate the quantitative and qualitative results.
% Then we provide ablation studies of our proposed method to show the effects of the components.
More video visualization is shown in the supplementary materials.

%%------------------------------------
%% ablation studies
\begin{table*}[h]
\renewcommand{\tabcolsep}{8pt}
\small
% \begin{adjustbox}{width=\columnwidth,center}
\centering
\begin{center}
\begin{tabular}{lccccccc}
\hline
\multicolumn{1}{c}{\multirow{2}{*}{\textbf{Method}}} & \multicolumn{2}{c}{\textbf{Motion Quality}}  & \multicolumn{2}{c}{\textbf{Motion Diversity}}   & \multicolumn{1}{c}{\textbf{Motion-Music Corr}}       & \multicolumn{1}{c}{\textbf{Freezing}}  & \multicolumn{1}{c}{\textbf{User}}                \\ \cmidrule(lr){2-3} \cmidrule(lr){4-5} \cmidrule(lr){6-6}
\multicolumn{1}{c}{}
&$\textrm{FID}_k \downarrow$ &$\textrm{FID}_g \downarrow$ & $\textrm{Dist}_k \uparrow$ & $\textrm{Dist}_g \uparrow$         & BeatAlign $\uparrow$ & \textbf{Rate}  & \textbf{Study} \\ \hline
GT & - & - & 9.06 & 7.31 & 0.292 & 18.7 \%  & 42\%\\
\hline
% Li et al
Li~\etal & 86.43 & 43.46 & 6.85 &3.32 & 0.232 & 59.0\%  & 98\%\\
% \hline
% % DanceNet
% DanceNet~\cite{} & 69.18 & 25.49 &2.86 &2.85 & 0.232 & 46.8\% \\
% \hline
% DanceRevolution
Revolution & 73.42 & 25.92 & 3.52 & 4.87 &0.220 & -  & 93\%\\
% \hline
% FACT
FACT & 35.35 & 12.40 & 5.94 & 5.30 & 0.241 & 39.0\%  & 90\%\\
% \hline
% Bailando
Bailando & 28.16 & \textbf{9.62} & 7.83 & 6.34 & 0.233 & 35.0\%   & 81\%\\
% \hline
% Sun et al
Sun~\etal & \textbf{25.96} & 13.42 & 7.68 & {6.59} & 0.249 & 29.6\%  & 75\%\\
EDGE & 40.25 & 15.45 & 11.28  & 6.36  & \textbf{0.27}  & \textbf{17.7} \%    &60\% \\
\hline
Ours &27.35 &12.78 & \textbf{12.01} &\textbf{7.29} & 0.265 & 24.2 \% & - \\
\hline

\end{tabular}
\end{center}
% \end{adjustbox}
\vspace{-10pt}
\caption{Comparison to the state-of-the-art methods on the AIST++ test set~\cite{FACT}.}
\label{tab: comparisons}
\end{table*}
%%------------------------------------

%%------------------------------------
%% ablation studies
\begin{table*}[h]
\renewcommand{\tabcolsep}{8pt}
\small
% \begin{adjustbox}{width=\columnwidth,center}
\centering
\begin{center}
\begin{tabular}{lcccccccccc}
\hline
\multicolumn{1}{c}{\multirow{2}{*}{\textbf{Method}}} &\multicolumn{4}{c}{\textbf{Ablations}} & \multicolumn{2}{c}{\textbf{Motion Quality}}  & \multicolumn{2}{c}{\textbf{Motion Diversity}}   & \multicolumn{1}{c}{\textbf{Motion-Music Corr}}       & \multicolumn{1}{c}{\textbf{Freezing}}               \\ \cmidrule(lr){2-5} \cmidrule(lr){6-7} \cmidrule(lr){8-9} \cmidrule(lr){10-10}
\multicolumn{1}{c}{} & M & G & S & I
&$\textrm{FID}_k \downarrow$ &$\textrm{FID}_g \downarrow$ & $\textrm{Dist}_k \uparrow$ & $\textrm{Dist}_g \uparrow$         & BeatAlign $\uparrow$ & \textbf{Rate} \\ \hline
\multirow{2}{*}{Ours} & & & & &35.6 &15.03 & 6.11 &5.01 & 0.229 & 30.5\%  \\
 & $\checkmark$& & &  &28.13 &11.35 & 8.09 &5.78 & 0.229 & 26.6\%  \\
 & $\checkmark$&&$\checkmark$  & &\textbf{23.43} &12.69 & 8.18 &6.03 & 0.233 & 28.1\%  \\
 & $\checkmark$&$\checkmark$ & & &23.58 & \textbf{9.22} & 7.10 &6.17 & 0.232 & 27.7\%  \\
 & $\checkmark$&$\checkmark$ &$\checkmark$ & &{24.32} &{11.75} & {8.89} &6.20 & 0.235 & 27.5\%  \\
 & $\checkmark$&$\checkmark$ &$\checkmark$ & $\checkmark$ &27.35 &12.78 & \textbf{12.01} &\textbf{7.29} & \textbf{0.265} & \textbf{24.2} \%  \\
\hline

\end{tabular}
\end{center}
% \end{adjustbox}
\vspace{-10pt}
\caption{Ablation studies of our proposed method. Examining the impact of Motion Perceptual Losses (M), Global-Trajectory Modulation Layer (G), Shared Embedding Function (S), and Mutual Information Minimization (I).}
\label{tab: ablation}
\end{table*}
%%------------------------------------

%-------------------------------------------------------------------------
\subsection{Experimental Settings}
% refer to FACT
\noindent
\textbf{Dataset.}
Our experiments use the AIST++ dataset~\cite{FACT}, with 992 paired dance motions and music from 10 genres. The motions are in SMPL format~\cite{loper2015smpl}, at 60 FPS, and the music ranges from 80 to 135 BPM. We train on 952 sequences and test on 40.

\noindent
\textbf{Training and Inference Details.}
We used a Transformer decoder with 4 attention heads and 512-dim hidden representations as our backbone network.
% The input vector consisted of a 240-frame (4 seconds) music sequence, a 120-frame (2 seconds) past motion sequence, and a 20-frame future motion sequence, each with its own position encoding function. 
% We use a Conv1d layer to extract the motion features and a linear layer to encode music.
Our experiments were trained with 126 batch size and a learning rate of 1e-4 on an A100 GPU for 2 days, with 1,000 diffusion steps.
During inference, we generated long-term dance in an auto-regressive manner, using past motions as future motions.
The global root trajectory for future motion is the noised global root trajectory in the diffusion process.

% \medskip
\noindent
\textbf{Music and Motion Representations.}
The music representations are extracted by the publicly available toolbox \textit{librosa}~\cite{librosa}, including \textit{mel frequency cepstral coefficients (MFCC), MFCC delta, constant-Q chromagram, tempogram, and onset strength}.
% The music representations are 438-dim in total.
% % pose features
% Our framework can accept motion representation in either locations, rotations, \etc.
We use the 6-DOF rotation representation~\cite{zhou2019continuity_6drotation} for each joint and a root translation, concatenating with joint positions, velocities, and foot contact binary labels.

% diffusion .. steps

\begin{figure}[t]
\begin{center}
   \includegraphics[width=0.96\linewidth]{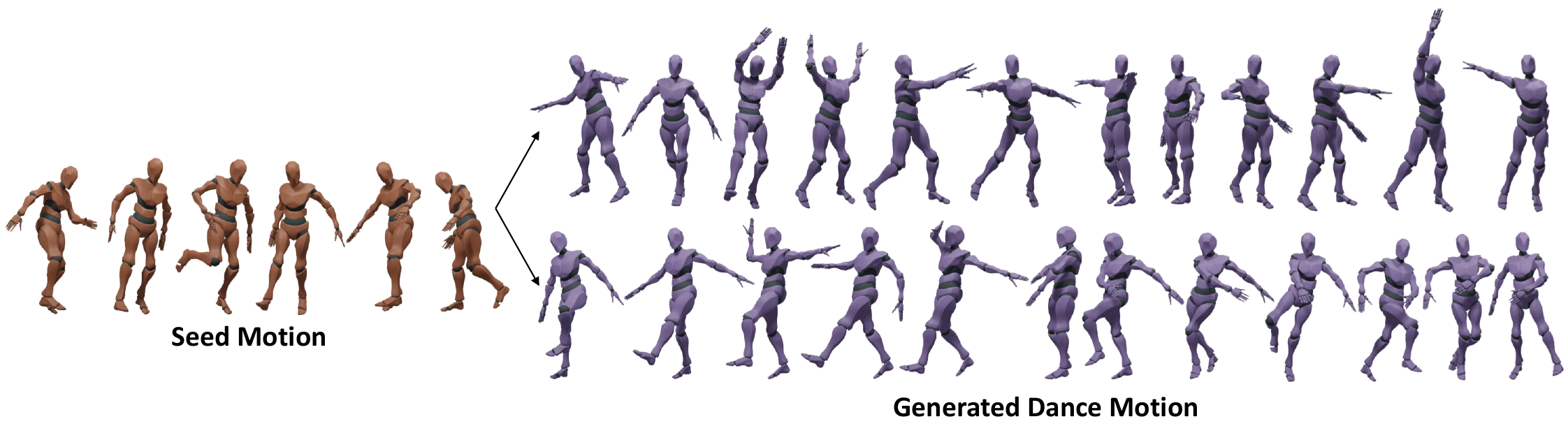}
\end{center}
\vspace{-10pt}
   \caption{Diverse Generation. Here we visualize two different dance motions generated from the same seed motion and music. It demonstrates the effectiveness of our method in generating diverse and high-fidelity dance motions.}
\vspace{-10pt}
\label{fig:seed_cmp}
\end{figure}

%-------------------------------------------------------------------------
\subsection{Comparison to the State-of-the-arts}
% 
% In this section, we evaluate our proposed method on different aspects, including 1) motion quality, 2) motion diversity, 3) motion-music correlation, and 4) long-term dance generation.
We test the 40 pieces of music in the test set of the AIST++ and generate dance sequences with a length of 20 seconds ($1,200$ frames).
We compare our method to Li~\etal\cite{li2020learning}, Revolution~\cite{dance_revolution}, FACT~\cite{FACT}, Bailando~\cite{bailando}, Sun~\etal\cite{you_never_stop} and EDGE~\cite{tseng2022edge}. 
% has not made their codes available and their reported experiments involve a different setting (i.e. 5-second dance generation in 30 FPS), we are not making comparisons with it.

%-------------------------------------------------------------------------
\subsubsection{Motion Quality}

Following the prior works~\cite{bailando,FACT}, we evaluate the motion quality by calculating the distance between motion features of the generated and the ground truth motions using Frechet Inception Distance (\textbf{FID})~\cite{heusel2017gans}.
We extract the following motion features implemented in \textit{fairmotion}~\cite{gopinath2020fairmotion}: 
(1) kinematic features (\lq\textit{k}\lq)~\cite{onuma2008fmdistance} that represent the velocity and acceleration of motions. It reflects the physical characteristics of dance.
(2) geometric features (\lq\textit{k}\lq)~\cite{muller2005efficient} that express geometric relation between specific body points.
It is based on multiple pre-defined templates of movements, which reflects the quality of choreography.
In Table~\ref{tab: comparisons}, it is demonstrated that our proposed method 
has comparable FID in kinematic and geometric features with other methods, showing that our proposed model has effectively captured the distribution of plausible motions.
% , achieving the best motion quality with a $\textrm{FID}_k$ score of 24.32 and a $\textrm{FID}_g$ score of 11.75, compared to Sun~\etal's scores of 25.96 and 13.42, respectively.
Despite obtaining the lowest score in terms of $\textrm{FID}_g$, Bailando's~\cite{bailando} model predicts joint positions instead of rotations. As a consequence, their predictions may result in non-uniform bone lengths caused by forcedly aligning with the target joint positions.
EDGE~\cite{tseng2022edge} argued that FID is not an unreliable metric for evaluation as the score does not improve even when the visual performance worsens.
% These experimental results suggest that our proposed conditional diffusion model has effectively captured the distribution of plausible motions.

%-------------------------------------------------------------------------
\subsubsection{Motion Diversity}
We compute the average Euclidean distance in the feature space to measure the diversity following~\cite{bailando,FACT}.
The motion diversities in the kinematic feature space and geometric feature space are denoted as $\textrm{\textbf{Dist}}_\textit{k}$ and $\textrm{\textbf{Dist}}_\textit{g}$, respectively.
It can be seen in Table~\ref{tab: comparisons} that our proposed method obtains the highest score (12.01) in $\textrm{{Dist}}_\textit{k}$ and $\textrm{{Dist}}_\textit{g}$ (7.29) in $\textrm{{Dist}}_\textit{g}$, which is in close proximity to the ground truth score.
% Moreover, the score of $\textrm{{Dist}}_\textit{g}$ (7.29) is comparable to EDGE ().
These findings establish the effectiveness of the proposed diffusion model in generating dance motions that are diverse and possess high fidelity. Some visualization results are shown in Figure~\ref{fig:seed_cmp}, where diverse motions are generated from the same music and seed motion.
% More video examples can be seen in our supplementary materials.

%-------------------------------------------------------------------------
\subsubsection{Motion-Music Correlation}
To evaluate the motion-music correlation, we measure the similarity between the kinematic beats and music beats.
The music beats are extracted using \textit{librosa}~\cite{librosa}, and the kinematic beats are the local minima of kinetic velocity.
To quantify the results, we compute the \textbf{Beat Alignment Score} (BeatAlign) following~\cite{FACT}.
% beat align score
The Beat Alignment Score is defined as the average distance between each kinematic beat and its nearest music beat.
% :
% \begin{equation}
% \footnotesize
%     \textrm{BeatAlign} = \frac{1}{|B^m|} \sum_{t^m\in B^m} \exp{(\frac{-\min_{t^d \in B^d} \|t^d - t^m\|^2}{2\sigma^2})} \textrm{ ,}
% \end{equation}
% where $B^m = \{t^m\}$ and $B^d = \{t^d\}$ are the music and kinematic beats, respectively.
% We set the normalized parameter $\sigma = 3$ as the FPS of motions in our experiment is 60.
As indicated in Table~\ref{tab: comparisons}, all approaches yield comparable beat alignment scores. We argue that achieving the highest score is not imperative for producing top-notch dance sequences. This is because in dance choreography, while it is expected that each dance step coincides with a specific musical beat, it is not obligatory for every musical beat to align with a dance step.

% beat hit rate (not sure to have it or not)

%-------------------------------------------------------------------------
\subsubsection{Long-Term Dance Generation}

% freezing rate (refer to 'you never stop')
The freezing problem is evident in long-term dance sequences.
To evaluate the quality of long-term dance generation, We calculate the \textbf{Freezing Rate} of each sequence following~\cite{you_never_stop}.
We divide a sequence into multiple non-overlapping sub-sequences of 60 frames.
For each sub-sequence, we calculate the average values of temporal differences of the pose $\Delta_{pose}$ and the translation $\Delta_{trans}$.
If $\Delta_{pose} \leq \tau_{pose}$ and $\Delta_{trans} \leq \tau_{trans}$ where $\tau_{pose}$ and $\tau_{trans}$ are predefined thresholds, the sub-sequence is regarded as a freezing sequence.
And the Freezing Rate measures the proportion of freezing sub-sequences in the generated motions.
Table~\ref{tab: comparisons} illustrates that EDGE has the lowest freezing rate, whereas our proposed method performs similarly. Despite EDGE achieving the lowest rate, it treats each motion segment independently, resulting in abrupt transitions, as illustrated in Figure~\ref{fig:frontpage}.
Compared to other methods conditioned on past motions, our proposed method effectively addresses the freezing problem, demostrating the effectiveness of our proposed mutual information minimization regularizer in balancing diversity and time coherency.

% % 15 seconds FID (long term)
% \TODO{15 seconds FID curve}
% refer to EDGE and long-term

% %-------------------------------------------------------------------------
% \subsubsection{Physical Foot Contact Score}
% We follow EDGE~\cite{tseng2022edge} to evaluate the physical plausibility of foot contact.
% It is a physically-inspired metric based on two observations: 1) 

% \begin{align}
% \textrm{PFC} &= \frac{1}{N \cdot \max_{1 \leq i \leq N} \| \bar{\Vec{\alpha}}^i_{\textrm{COM}} \| }\sum^{N}_{i=1} s^i \textrm{ ,} \\
% \intertext{where}  
% \bar{\Vec{\alpha}}^j_{\textrm{COM}} 
% &= (\Vec{\alpha}^j_{\textrm{COM}, x}, \; \Vec{\alpha}^j_{\textrm{COM}, y}, \; \max(\Vec{\alpha}^j_{\textrm{COM}, x}, 0))^T     \textrm{ ,} \\
% s^i &= \| \bar{\Vec{\alpha}}^i_{\textrm{COM}}  \|  \cdot \| \Vec{v}^i_{\textrm{Left Foot}} \| \cdot \| \Vec{v}^i_{\textrm{Right Foot}} \| \textrm{ .}
% \end{align}

% $\Vec{\alpha}$ is the acceleration, $\Vec{v}$ is the velocity and $i$ is the frame index.

%-------------------------------------------------------------------------
\subsubsection{User Study}

We conducted a user study to assess the motion-music relationship of our method and other methods. 25 participants watched 24 pairs of videos, each containing two dances generated by our method and other methods. Participants were asked to evaluate which video had better dancing. The results showed that our method outperformed FACT, Li et al., and Revolution in more than 90\% of cases due to these methods' freezing problems. Our method also outperformed Bailando in 81\% of cases due to the varying lengths of bones generated by Bailando. In high motion diversity, our method outperformed Sun et al. in 75\% of cases. Our method outperformed EDGE in 60\% of cases due to its better coherence in motions and style. However, 42\% of the ground truth dance motion was determined to be better than our generated dance motion. 
Our method still had less fidelity compared to the ground truth motions.

%-------------------------------------------------------------------------
\subsection{Ablation Studies}
We conduct ablation studies to verify the importance of each component in our proposed method.
% The results are shown in Table~\ref{tab: ablation}.

\medskip
\noindent
\textbf{Effect of Motion Perceptual Losses.}
Table~\ref{tab: ablation} shows that integrating motion perceptual loss by considering human motion cues in positioning and velocity results in superior motion quality, with reduced artifacts, and glitches. Moreover, the adoption of motion perceptual loss, particularly velocity loss, decreases the freezing rate by minimizing the deviation between predicted and ground-truth velocities, ensuring consistency with human perception.

\medskip
\noindent
\textbf{Effect of Shared Embedding.}
In Table~\ref{tab: ablation}, we can see that we could obtain the lowest FID in kinematic features with the use of the shared embedding function. It is noteworthy that kinematic features are particularly sensitive to noisy motions. Thus, it shows that the shared embedding function is effective in aligning the distributions of real and noisy motions, which in turn facilitates the removal of noise in motion, leading to a reduction in the FID score. 

\begin{figure}[t]
\begin{center}
   \includegraphics[width=0.98\linewidth]{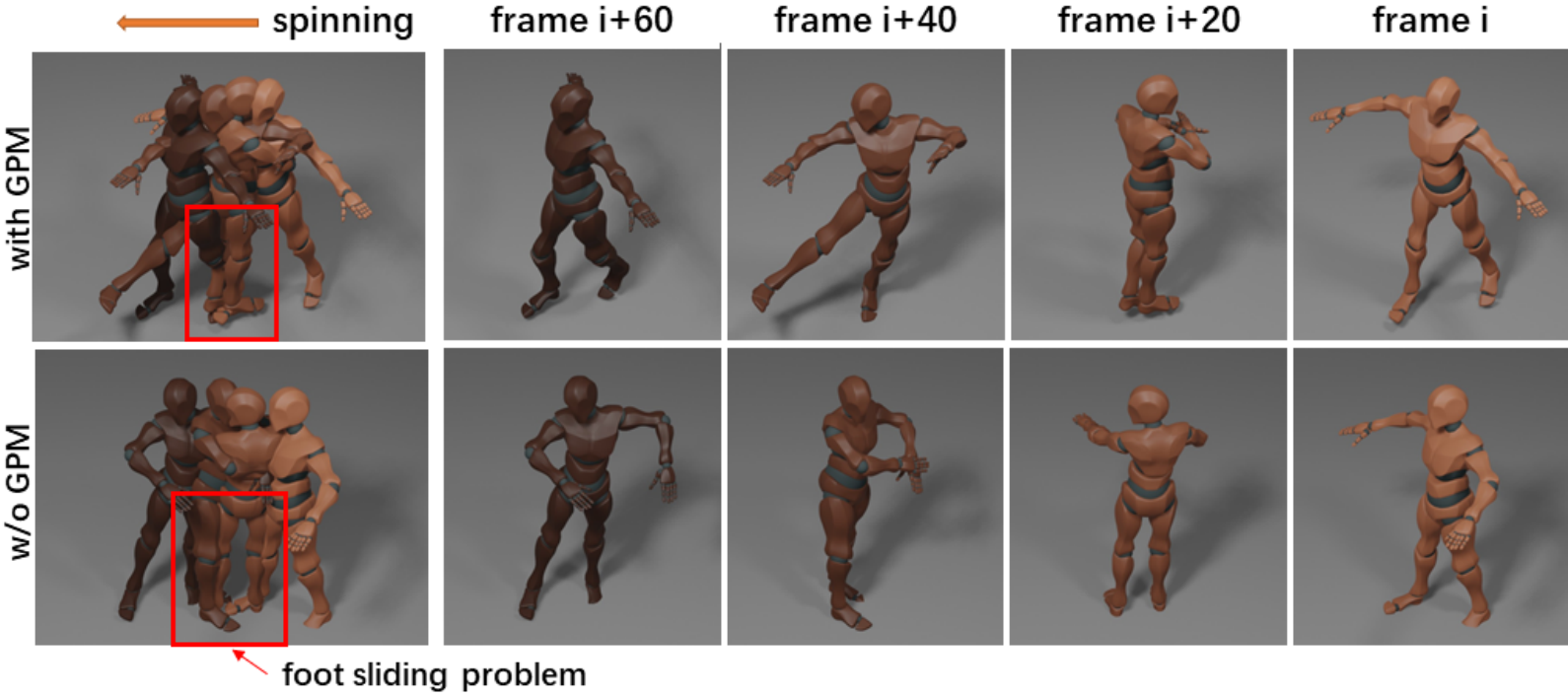}
\end{center}
\vspace{-10pt}
   \caption{Effect of global-trajectory modulation layer. The upper row shows that with the GTM layer, the right foot can remain in the same position without sliding.}
\vspace{-10pt}
\label{fig: foot contact}
\end{figure}

\medskip
\noindent
\textbf{Effect of Global-Trajectory Modulation Layer.}
Table~\ref{tab: ablation} shows that the inclusion of the global-trajectory modulation layer leads to lower FID scores for both kinematic and geometric features. This indicates that the global-trajectory modulation layer effectively enhances the refinement of global root trajectory, resulting in more realistic generated motions. However, this improvement in fidelity incurs a trade-off with decreased diversity. 
Therefore, the integration of the global-trajectory modulation layer can be seen as a promising approach for generating high-fidelity motion sequences, but it may require further optimization to maintain diversity in the generated outputs.
Combing it with other components in our proposed method, we achieve the overall trade-off of diversity and high fidelity.
Figure~\ref{fig: foot contact} illustrates that our proposed GTM layer enables the right foot to maintain a fixed position on the ground, whereas the right foot in the lower row slides from right to left during spinning.

\medskip
\noindent
\textbf{Effect of Mutual Information Minimization.}
Table~\ref{tab: ablation} demonstrates the significant impact of incorporating mutual information minimization into our model. The inclusion of this technique led to a substantial increase in the diversity of $\textrm{{Dist}}_\textit{g}$ and $\textrm{{Dist}}_\textit{g}$, resulting in the generation of a wider range of motions that are not present in models without this feature. Additional visual examples are provided in the supplementary materials. Importantly, the freezing rate was notably reduced from 27.5\% to 24.2\%, underscoring the effectiveness of this approach in mitigating over-reliance on past motions. These results collectively highlight the efficacy of our proposed technique in enhancing diversity and reducing the freezing rate in generated dance sequences.

%------------------------------------------------------------------------
\section{Conclusions}
We presented LongDanceDiff, a transformer-based diffusion model for generating dance sequences conditioned on music and past motions. The proposed model has been extensively evaluated using user studies and several metrics and has achieved state-of-the-art results. Our generated samples show that LongDanceDiff is capable of creating long and diverse dance sequences with high temporal coherency, which highlights the effectiveness of the mutual information minimization regularizer. To address the issue of foot sliding and improve the overall quality of the generated dance motions, we have also incorporated spatial constraints to the diffusion model via a global-trajectory modulation layer and motion perpetual losses. These results demonstrate the potential of our approach for generating high-quality dance sequences with diverse and coherent motions.
% LongDanceDiff demonstrates the powerful ability of the diffusion model in the field of music-conditioned dance generation.

\bibliography{egbib}

\end{document}